# Development of an Ontology to Assist the Modeling of Accident Scenarii "Application on Railroad Transport "

Ahmed Maalel, Habib Hadj mabrouk, Lassad Mejri and Henda Hajjami Ben Ghezela
**Abstract**— In a world where communication and information sharing are at the heart of our business, the terminology needs are most pressing. It has become imperative to identify the terms used and defined in a consensual and coherent way while preserving linguistic diversity. To streamline and strengthen the process of acquisition, representation and exploitation of scenarii of train accidents, it is necessary to harmonize and standardize the terminology used by players in the security field. The research aims to significantly improve analytical activities and operations of the various safety studies, by tracking the error in system, hardware, software and human. This paper presents the contribution of ontology to modeling scenarii for rail accidents through a knowledge model based on a generic ontology and domain ontology. After a detailed presentation of the state of the art material, this article presents the first results of the developed model.

**Index Terms**— Human safety, Knowledge acquisition, Ontology design, Transportation.
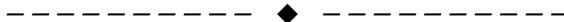

## 1. Introduction

Negative impact, the terrible cost of accidents, the occurrence of future disasters despite the advances in technology are the basis for the establishment of a process of feedback (Rex) as an essential means to promote the necessary improvement in security. The concept of "feedback" on transportation security is defined differently by authors and fields.

The part common to all definitions is the interest of learning from an experience to avoid reproduction. The Rex is a dynamic process of collection, storage, analysis and exploitation of data relating to breach security situations (accident or incident). It consists in an analytical study of different causal factors involved in the genesis of incidents or accidents. Rex allows a better understanding of the mechanisms leading to events of insecurity. Its purpose is to benefit from the lessons of past experience to improve the level of security by implementing preventive measures and correctives to prevent accidents and incidents [1].

The methods of acquiring knowledge, applied to the field of security analysis, have shown their interests to extract and formalize the historical knowledge of security analysis (mainly accident scenarii) and its limits in terms of extraction of gait analysis and expert evaluation of safety, especially based on intuition and imagination. Generally, current methods of acquiring knowledge have been developed for well-structured problems. They do not address the specifics of the multi-expertise and coexistence of diverse knowledge and do not allow access to the subjective and intuitive knowledge related to a highly scalable and field unbounded as is that of security and certification of rail transport systems. Indeed, the acquisition of knowledge has encountered the difficulty of extracting expertise raised at each stage of the process of analysis and safety assessment. This difficulty stems from the complexity of the expertise that naturally encourages the experts to give their expertise through significant examples or scenarii of accidents experienced on automated transit systems already certified.

## 2. System ACASYA
### 2.1 Presentation

In contrast to diagnostic support systems, "ACASYA" can be seen as a tool for the prevention of accidents at the stage of system design. The objective of this tool is, firstly to assess the completeness and consistency of accident scenarii proposed by manufacturers and secondly to contribute to the generation of new scenarii that could help experts certification. Specifically, the tool "ACASYA" can help experts of INRETS, even the builder and the client, particularly in the assessment phase of the completeness of functional assays Security (AFS). The work performed under the two theories [2], [3] LAMIH at the University of Valenciennes and pursued INRETS ESTAS-enabled:

1. To establish a base that includes 70 accident scenarii;
2. Develop a formalism scenarii (static and dynamic description);
3. To develop a model of feasibility ACASYA structured around three modules Fig. 1 :

---
- Maalel A. is with RAID Labs. (National School of Computer Science of Manouba). E-mail: ahmed.maalel@ensi.rnu.tn
- Hadj Mabrouk H. is with French Institute of Science and Technology of Transport, Planning and Networks. E-mail: mabrouk@inrets.fr
- Mejri L. is with RIADI Labs.( National School of Computer Science of Manouba). E-mail: mejrilassad@yahoo.fr.
- Ben Ghezela H. is Director of the RIADI labs. ( National School of Computer Science of Manouba ). E-mail: hhbg.hhbg@gnet.tn
1

- CLASCA is a model of inductive learning system for classification of scenarii;
- EVALSCA is a model of expert system to aid scenario assessment. This model, developed around the learning system of rules CHARADE J.-G. Ganascia (LIP6) aims to suggest to experts possible failures not considered in the safety analysis;
- GENESCA is a model system for assisting the generation of scenarii. This feasibility model does not systematically generate new scenarii relevant and usable, but only embryos of scenarii that stimulate the imagination of experts in the formulation of scenarii of accidents.

## 2.2. Model representation of accident scenarii

### 2.2.1. Accident scenario

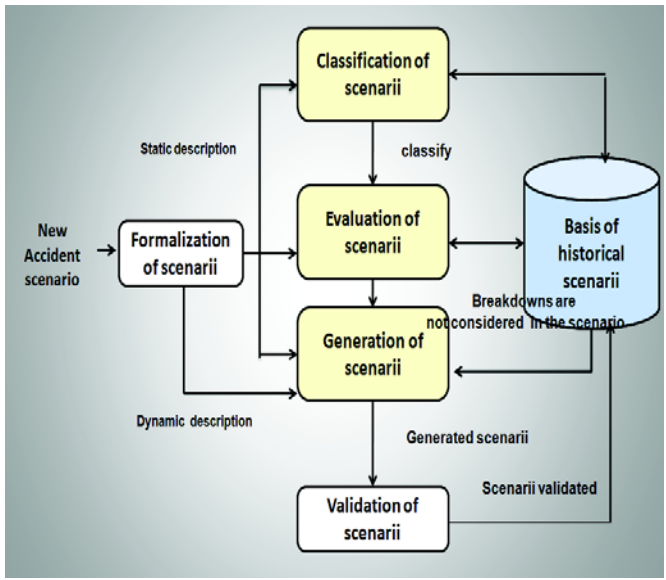

Fig. 1. Organisation functional model of the system ACASYA [2].

An accident scenario describes a combination of circumstances that can lead to undesirable and even dangerous situations. It is characterized by a context and a set of events and parameters. Phase Micro-extraction of knowledge has led to the shaping of a particular model based on the identification of eight parameters describing an accident scenario [2] , [4].

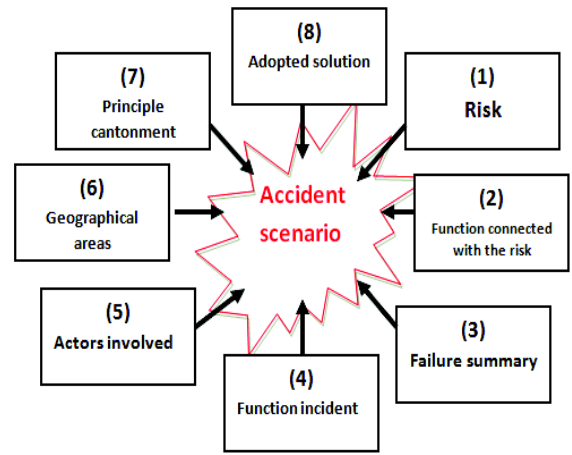

Fig. 2. Characteristic parameters of an accident scenario

### 2.2.2. Static description of an accident scenarii

The static description lists the main parameters of an accident scenario in a "fact sheet" prepared as a table of data (Attributes / Values). The attributes correspond to the eight parameters and each attribute is associated to a list of possible values.

This information has subsequently been used as a basic form for the acquisition scenarii, it is exhaustive and can approach the study of any type of automated transportation system (VAL MAGGALY POMA). In summary, the static description of a scenario has led to the definition of a first language for describing examples of scenarii. This is a classical representation by attribute-value pairs. This language of expression is close to the language of the expert certifier and has the advantage of being compatible with the structure of historical data security.

TABLE 1
Fact sheet parameters of an accidents scenarii

| CHARACTERISTICS PARAMETERS OF AN ACCIDENT SCENARIO | | | Chosen Values (X) | Key Concepts (*) |
|---|---|---|---|---|
| | List of attributes | List of possible values | | |
| 1 | geographical Principle | Fixed Canton | | |
| | | moving Canton | x | * |
| 2 | Risks | Collision | x | * |
| | | Derailment | | |
| | | Poorly controlled Emergency Evacuation | | |
| | | Drop in the vehicle | | |
| | | Fall on the way | | |
| | | Entrainement de l'individu | | |
| | | Evacuation | | |
| | | Hour of closing the doors | | |
| 3 | Functions in connection with the risk | Management of automatic driving | x | * |
| | | Localisation des trains | | |
| | | Control Input / Output | | |
| | | Tracking trains | | |
| | | Travel Direction Management | x | * |
| | | Speed setpoint | | |
| | | Management of trains Stops | | |
| | | Security Pier Lane | | |
| | | Authorization CI / HT | | |
| | | Redundancy switching | | |
| | | Initialization | | |
| | | Manual Driving | | |
| | | Alarm Management | | |
| | | Evacuation | | |
| | | Docking | | |
| | | Routes Protection | x | * |
| | | Traction / Braking | | |
| 4 | Geographical areas specified in scenario | Terminus | x | * |
| | | Session | | |
| | | Way | | |
| | | Injection zone of oar | x | * |
| | | Segment limit | | |
| 5 | Actors involved in scenario | Number of trains | 2 | * |
| | | Operator at CCPs | | |
| | | Mobile operator | | |
| | | PA with redundancy | | |
| | | PA without redundancy | x | * |
| 6 | Incidental Functions | Route management | x | * |
| | | Traffic regulation | | |
| | | Instructions (consistency, vigilance) | | |
| | | Communication (transmission) | | |
| 7 | outlined Outages caused by the scenario | OO26 | Element and target in opposite direction | x | * |
| 8 | Interim solutions | IS | | | |
| | Optimal solutions | OS15 | Compare the meaning of the element target meaning | x | |

## 2.2.3. Dynamic description of an accident scenario

The dynamic description is based on the use of two modes of representation of a scenario: Petri and Table Sequencing of marking.

The Petri net tool, formal specification and description of the systems have been used to model the behavior of the equipment or the transport system challenged by the script. To accommodate the different elements of the scenario modeled, Petri dissociates three aspects:

The first aspect describes the system's external environment (for example, the passage of a train from one station to another).

The second aspect takes into account the internal environment, i.e. automatic as autopilots, alarms…

The third describes the interface between these two types of environment that ensures the exchange of information.

Modeling a scenario of an accident by a Petri then used simulation to examine all outstanding situations deemed dangerous or undesirable in terms of security. In practice, the Petri prepared for a scenario allows a more comprehensive study of various possible situations of insecurity. Each of these situations is described as a table called "sequencing of the CE marking.

Table Sequencing of Marking: this chronology of events preceding the partial or total alteration of the security. [2], [3] It is characterized by three basic concepts:
- An initial state (initial marking)
- A chronology of situations,
- A critical situation (that is somehow a pre-accident because it announces the inevitable accident).

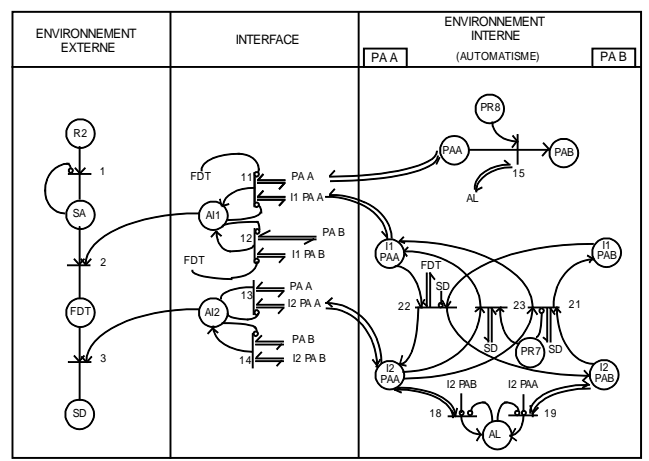

Fig. 3. Modeling the scenarii by R. Petri [2], [3].

## 3. Limits of works

The study ACASYA raised three main limitations:

- Limits on the model representation of accident scenarii: a sheet of 8 parameters: a textual description is not sufficient to formalize an accident scenario, we lose the richness and semantic aspects of knowledge. All descriptions (static / dynamic / text / graphics and) were not all used by ACASYA. Indeed, classification, evaluation and generation of accident scenarii have not used these various forms of representation jointly. It would be beneficial to make the most of the 4 forms of additional representations for the concept of accident scenario.
- Not taking into account human factors although human error is responsible for 65% of accidents in railway transport [6]-[8].
- Limits on exploitation knowledge archived. Indeed the evaluation scenarii in ACASYA relied solely on a particular attribute of the description: that of static failures summarized. Although this attribute is important, it would be better to extend further the assessment base. In addition, the scenario generation has neglected the dynamic description of a scenario (the Petri network that illustrates the dynamics of sequencing an accident in space and time).

Our work in this paper aims to provide a partial answer to the first limitation mentioned. Indeed, despite the interest representation model accident scenarii adopted in ACASYA, it seems necessary to standardize and harmonize the representation formalism and the terminology used by the players area. This terminology is suffering from the same view of the incompleteness and inconsistency of the model representations used, and it could be improved. In this context, ontology is well- suited to understand and improve significantly the representation formalism.

In addition, the formalism adopted to represent these accident scenarii is limited to a purely technical aspect. However, an accident can be generated only through a Human/ Machine poor cooperation and / or environment, therefore, it seems obvious to be involved in modeling the human component scenarii.

Before presenting the first results of the representation model developed, we will first present ontologies. The following paragraph is devoted to the overall presentation of ontologies

## 4. Ontology

### 4.1. Presentation

Generally, an ontology is "the specification of a conceptualization of a knowledge domain" .This definition is based on two dimensions: An ontology is the conceptualization of a domain, i.e. tell a choice about how to describe a domain. It is also the specification of this conceptualization, that is to say its formal description. According to Natalya F. Noy and Deborah L. McGuinness [9] An ontology defines a common vocabulary for researchers who need to share information in a domain. It includes machine-readable definitions of basic concepts in this area and their relationships

Ontology has gotten a pivotal place in many areas: Artificial Intelligence in Semantic Web, software engineering to biomedical informatics, and now, information architecture is considered a form of knowledge representation. [10] Given the interest they represent, ontologies have been the subject of standardization. We can cite for example the ISO 21127 standard (ISO 21127, August 2002) entitled "Ontology needed to describe data concerning cultural heritage." This standard was published in 2006 following the definition of intangible cultural heritage conducted by the UNESCO. It describes in particular the metadata necessary for structuring ontology.

Ontology is generally used for reasoning about objects in the relevant field modeled after a certain body of knowledge within the field. So, ontology is itself a data model representing a set of concepts in a domain, and the relationship between these concepts [10].

### 4.2. Types of ontology

Depending on their use, there are five categories of ontology [11] in this case: generic, domain, application, and representation of method:

- Generic Ontology: describes general concepts, independent of a field or a particular problem. Examples of concepts of time, space and event.
- Domain Ontology: specifies a point of view on a particular area using vocabulary related to a domain of generic knowledge. The concepts of an ontology is often defined as a specialization of the concepts of generic ontology. A domain ontology consists of a description of the extension of the vocabulary domain, a typology and hierarchy or a lattice of classes.
- Application Ontology: Describes the structure of knowledge necessary to achieve a particular task [11]. It enables domain experts to use the same language as the application.
- Representation Ontology: defines a set of primitive concept representations of domain ontology and generic ontology.
- Method Ontology: describes the process of reasoning independently of a domain and a given implementation. It specifies the entities that fall under the resolution of a problem and provides definitions of concepts and relations used to specify a process of reasoning when performing a particular task.

## 5. Proposed approach

### 5.1. Conceptual Model

Very schematically, A formalization makes optimal ontology reuse, although reusability is confronted with usability. In other words, the more reusable an ontology is the more it loses in accuracy and therefore moving away from the application scenario and loses its usefulness. And an ontology is more specialized, the more it is close to the concerns of an application [12].

The ontology in our context, uses two levels of abstraction of knowledge that we can qualify by knowledge of surface and deep knowledge. The knowledge of surface are generally defined by what we call a generic ontology and the insights are defined by the domain ontology.

- A generic ontology: An ontology contains generic concepts such as Security: Background, Components dangerous, dangerous events, Causes of the accident, etc.

- A domain ontology: This ontology describes the field of rail transportation safety, related to the specific system, the human operator and the environment. Most of these concepts are specializations of other concepts of the generic ontology.

To represent the conceptual model, a choice was on the mind maps (mind maps) for their effectiveness in representing and visualizing hierarchies of concepts. They are also quite common and easy to handle (the publisher FreeMind 1 was used). However, as mind maps are not intended for the conceptualization of ontology, we have proposed some principles that operate the various elements of the publisher FreeMind (node, icon, comment, etc..) To model the elements of ontology (concept, property, authority, etc..) needed in the conceptualization (Fig. 4,5,6) [13]:

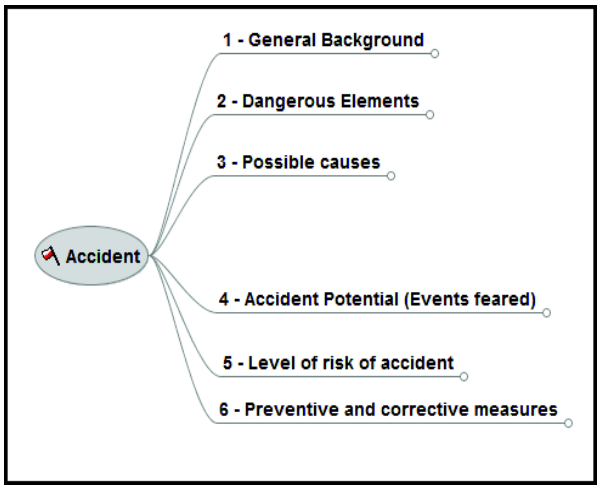

Fig. 4. The main classes of the generic ontology

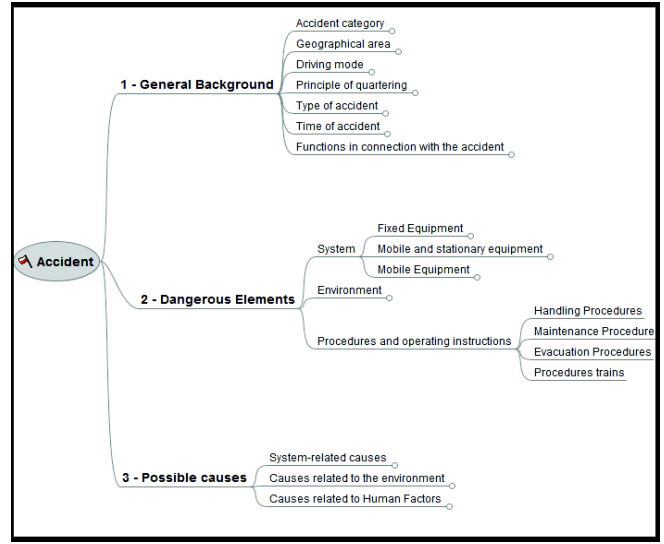

Fig. 5. The extracted hierarchy of classes in the developed ontology

An instance of a concept is represented by a node with an icon 1.

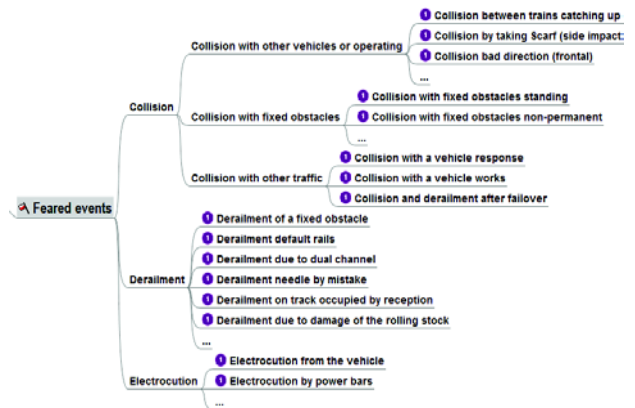

Fig. 6. Extract a few instances of concepts (Class: Event feared)

### 5.2. Construction of the ontology

Knowledge engineering involved in making a machine-readable ontologies. It is responsible for "modeling ontological" is to define the primitives of representation and meaning that will be used for the formal modeling of knowledge, a preliminary to formal modeling, that is to say, representation in formal language domain knowledge [14].

Although there are several construction methods of the ontology, we adopted the method of Stanford University, implemented with the publisher protected. We conducted our choice on the basis of two arguments described below:

1. The method of Stanford University is detailed and promotes ease of development of ontologies for both professionals and for beginners;

2. The ontology editor protege, is the most used. His popularity forced the editors of other ontologies (OntoStudio) to develop several gins that will do their ontologies. Protection is more flexible and better con-aunt by her community because it is an open source software (unlike Ontostudio example that requires a license);

Fig. 7. Shows the different classes of the developed ontology with the tool OwlVIZ integrated into the editor "the protégé" ;

## 5.3. Extracting knowledge from the ontology

We will in this pragraphe presented the first results of development of a model for querying the ontology developed.

The developed model is based on an WEB approach. Indeed, a team of German scholars has recently developed a tool to use XML files: where one file extensions obtained after the construction of an ontology is an RDF / XML file (actually a file XML). We chose this tool for the operation of our ontology. The tool eXist-DB is an open source management database built using XML technology. It allows to use the RDF file by exploiting the XQuery query language that supports. XQuery is a query language for XML files with syntax very similar to the SQL language.

Fig. 8. presents the functional architecture of the model developed that focuses on three main tasks: the acquisition, archiving and consultation for accidents;

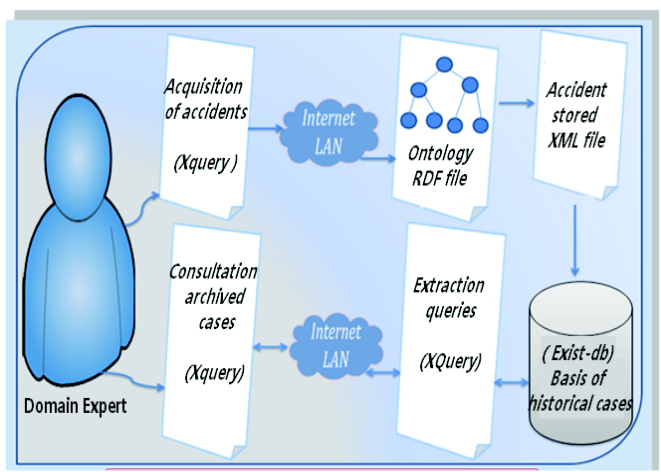

Fig. 8. Functional Architecture of the developed model

Fig. 9. presents the steps of acquiring a case. This is the first module of the system, which accesses the user expert once identified. The expert user can access the main classes, subclasses, including to instances extracted from the ontology. The expert user can select one or more instances possible (leaves) for each subclass.

This model allows to automatically associate an XML file for each case (accident scenario) acquired.

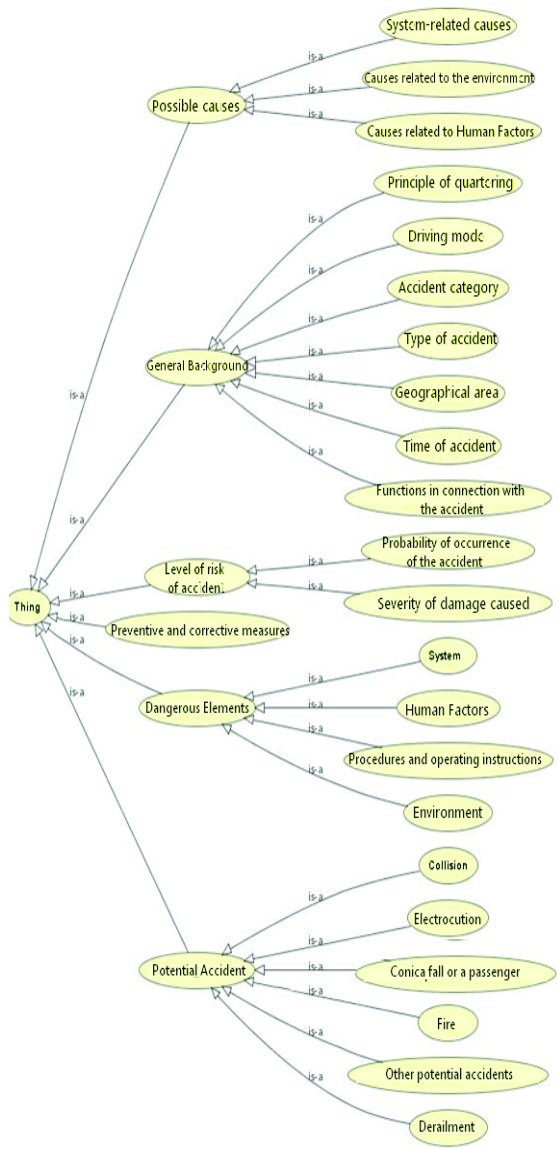

Fig. 7. Diagram of different classes of the ontology with Owl VIZ

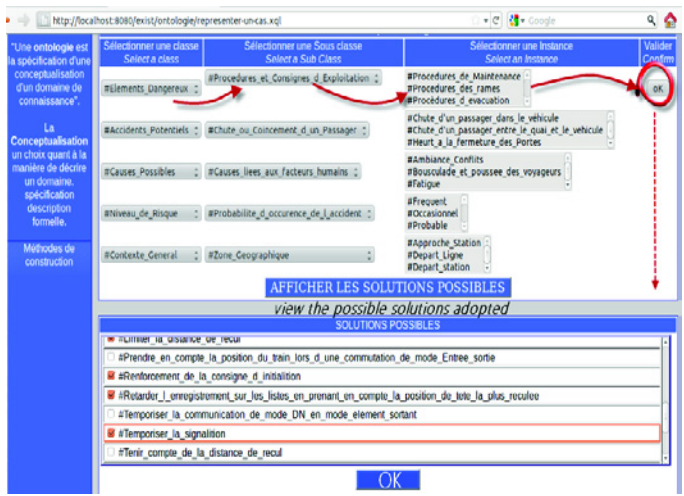

Fig. 9. Acquisition window of accidents

## 6 CONCLUSION

The treatment of an accident scenario requires passage through several major phases: an acquisition phase, a modeling phase, a phase of capitalization and operation phase. To date and to our knowledge there are no formal and systematic approaches to exploit the knowledge related to accident scenarii. The research is generally limited particularly in the context of rail transport to the phases of capitalization on the operational phase, it is generally limited to a purely statistical phase. At the end to improve and take advantage of this process of Rex, it is essential to develop a knowledge representation formalism most appropriate and most semantic end to prepare the ground for the next phase of operations and recommendations to propose and to identify know-how which can help stimulate and domain experts in their task of certification and accreditation. This paper has presented the first results of work on the subject

**Ahmed Maalel** received his Master of Science of Transportation and Logistics in the Higher Institute of Transport and Logistics of Sousse. He has published many papers in international conferences , received also the Paper Award in the SIIE2010 conference. He is currently a PhD student in the RIADI labs. in the National School of Computer Science of Manouba.

**Habib Hadj Mabrouk** is currently a researcher at the French Institute of Science and Technology of Transport, Planning and Networks. he received his accreditation to supervise research (HDR) in 1998. His research Interests links safety and risk analysis in the Railroad Transport, the return of experience Rex and human factors.

**Lassad Mejri** is currently Assistant Professor of Computer Sciences in the department of Computer Sciences at the Faculty of Science of Bizerte. he obtained his doctoral thesis from the Valenciennes University in 1995.

**Henda Ben Ghezala** is currently Professor of Computer Science in the department of Informatics at the National School of Computer Sciences of Tunis. She leads a Master degree in 'ICIS'. She is the president of University of Manouba. Her research interests lie in the areas of information modeling, databases, temporal data modeling, object-oriented analysis and design, requirements engineering and specially change engineering, method engineering. She is Director of the RIADI labs.